\documentclass{article}
\usepackage{cite}
\usepackage{amsmath,amssymb,amsfonts}
\usepackage{spconf}
\usepackage[colorlinks=true, allcolors=blue, urlcolor=purple]{hyperref}
\usepackage{algorithmic}
\usepackage{graphicx}
\usepackage{textcomp}
\usepackage{xcolor}
\usepackage{soul}
\usepackage{cuted}
\usepackage{setspace}

\usepackage{multirow}
\usepackage{booktabs}
\usepackage{diagbox}
\usepackage{amsmath,epsfig}
\newcolumntype{?}{!{\vrule width 1pt}}

\usepackage{colortbl}
\definecolor{Gray}{gray}{0.85}
\newcolumntype{a}{>{\columncolor{Gray}}c}

\def\BibTeX{{\rm B\kern-.05em{\sc i\kern-.025em b}\kern-.08em
    T\kern-.1667em\lower.7ex\hbox{E}\kern-.125emX}}

\newcommand\x{0.19}

\begin{document}
\title{ Weakly-supervised continual learning\\ for class-incremental segmentation
}

\name{Gaston Lenczner\textsuperscript{1,2}, Adrien Chan-Hon-Tong\textsuperscript{1}, Nicola Luminari\textsuperscript{2}, Bertrand Le Saux\textsuperscript{3}\thanks{Corresponding author: \href{mailto:gaston.lenczner@alteia.com}{gaston.lenczner@alteia.com}}} 
\address{\textsuperscript{1}ONERA/DTIS, Universit{é} Paris-Saclay, FR-91123 Palaiseau, France\\\textsuperscript{2}Alteia, FR-31400 Toulouse, France
\\\textsuperscript{2}ESA/ESRIN $\Phi$-lab,  I-00044 Frascati, Italy }

\maketitle
\begin{abstract}
Transfer learning is a powerful way to adapt existing deep learning models to new emerging use-cases in remote sensing. Starting from a neural network already trained for semantic segmentation, we propose to modify its label space to swiftly adapt it to new classes under weak supervision. To alleviate the background shift and the catastrophic forgetting problems inherent to this form of continual learning, we compare different regularization terms and leverage a pseudo-label strategy. We experimentally show the relevance of our approach on three public remote sensing datasets. Code is open-source and released in this repository: \hyperlink{https://github.com/alteia-ai/ICSS}{https://github.com/alteia-ai/ICSS}.
\end{abstract} 

\begin{figure}[ht]
  \begin{minipage}[t]{.44\linewidth}
  \centering\epsfig{figure=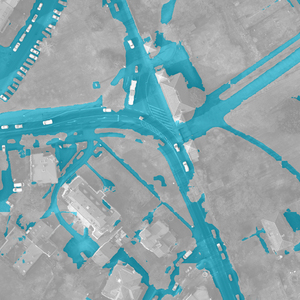, width=\linewidth}
    (a) Initial prediction
  \end{minipage} \hfill
  \begin{minipage}[t]{.44\linewidth}
  \centering\epsfig{figure=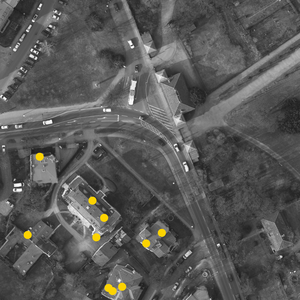,width=\linewidth}
    (b) New class annotations
  \end{minipage} \hfill
  
  \begin{minipage}[t]{.44\linewidth}
  \centering\epsfig{figure=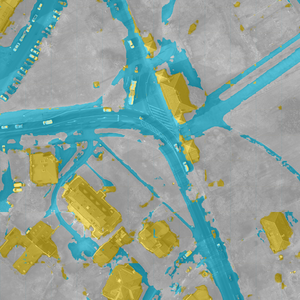,width=\linewidth}
    (c) New prediction
  \end{minipage} \hfill
    \begin{minipage}[t]{.44\linewidth}
  \centering\epsfig{figure=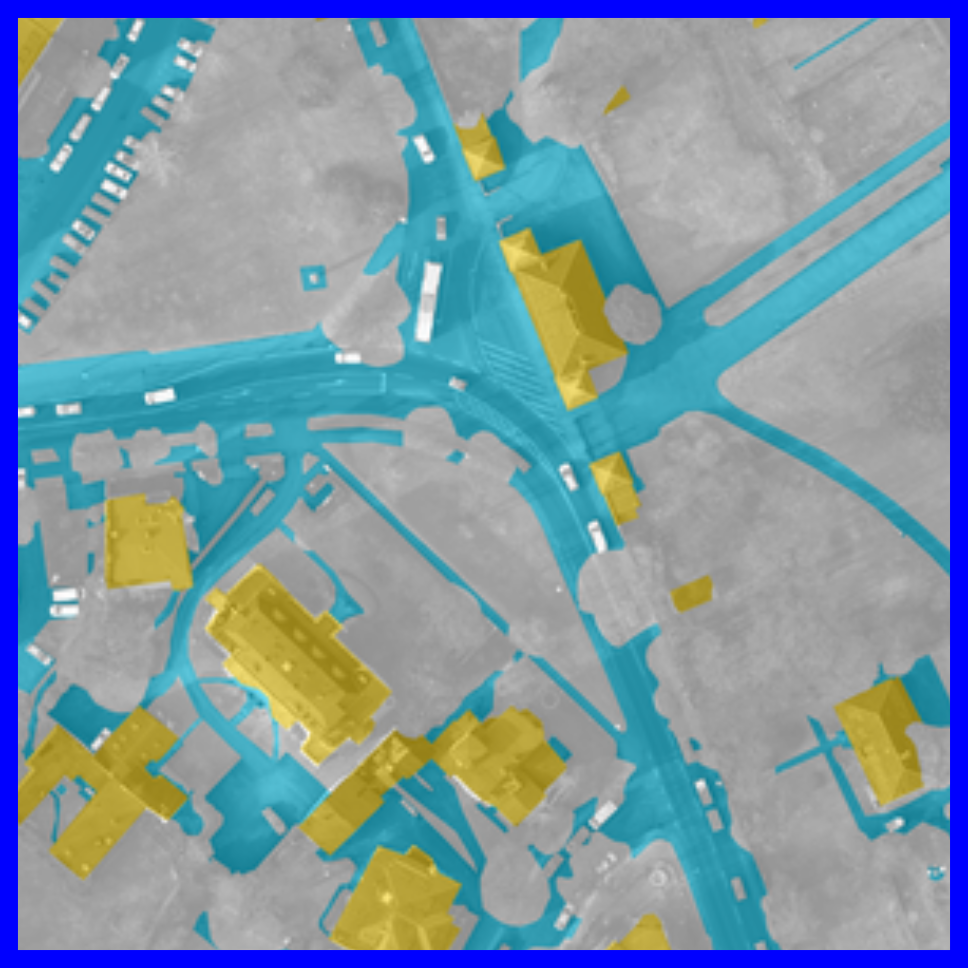,width=\linewidth}
    (d) Ground-truth
  \end{minipage} \hfill
  \caption{Add \textit{building} in the \{\textit{background}, \textit{road}\} label space with clicked annotations. Initial prediction (a) leads to new-class annotations (b) to collect the new prediction (c).} 
  \label{fig:visual}
 \end{figure}
 
 \section{Introduction}


Semantic segmentation, or pixel-wise classification of an image, is a challenging computer vision task with various remote sensing applications such as land-cover mapping, building recognition or change detection. Nowadays, deep neural networks (DNNs) are commonly applied for such purposes. They are usually trained once on a given training dataset and then deployed to automatically process potentially large volumes of data. 


This approach outcomes frozen algorithms that can lead to excellent outputs in ideal conditions. However, there are still many open questions, such as domain adaptation or limited training data, for which such a static approach is often not suitable. Another important issue of these algorithms is their lack of flexibility due to their frozen output space fixed before training.
Many works now use semi-supervised learning~\cite{castillo2021semi} and weakly supervised learning~\cite{hua2021semantic} to address limited training data issues and domain adaptation~\cite{lucas2021bayesian,wang2019weakly}. %
Another promising approach to handle these problems comes with interactive learning~\cite{lenczner2020interactive}, where the algorithm learns to adapt to user inputs. To improve the neural networks plasticity for semantic segmentation, recent works propose customized losses for class-incremental segmentation. 
 
Even though fine-tuning a deep network on a small set of samples of new classes seems quite risky due to both catastrophic forgetting and lack of convergence threats, we show  in this work that this is possible within the correct framework. Precisely, we combine weak and class-incremental segmentation works to propose to increment neural network label space using sparse clicked annotations, as showed by Figure~\ref{fig:visual}. This set-up is suited for the interactive learning of new classes since these annotations are particularly handy for an annotator~\cite{hua2021semantic}. Currently, it might be easier to ask an operator to annotate training data with the new segmentation classes to fully retrain the algorithm. Yet, the purpose of the proposed framework is precisely to be able to learn a new class on new data. This method is therefore suitable for use-cases where initial training data is unavailable due to privacy concerns or where accurate predictions are needed quickly on small datasets.

To this purpose, we first compare different regularizations to fully benefit from the sparse annotations. We also explore a pseudo-label strategy to alleviate catastrophic forgetting. To summarize, the contributions of this paper are: (1) A framework to increase the label space of a DNN for semantic segmentation with sparse annotations; (2) A pseudo-label strategy and a comparison of different regularizations for this weakly supervised fine-tuning; (3) An experimental validation of the proposed approach on three public remote sensing datasets.



\section{Related work}

\subsection{Semi and weak supervision in semantic segmentation}

When labels are scarce or flawed, the common training paradigm is to learn to perform the segmentation task with the available labels while leveraging unlabeled data to learn a better inner representation as support. In semantic segmentation, weak labels can take various forms like incomplete ground-truth maps or bounding boxes. Recently, different works in remote sensing explore these different aspects.
In~\cite{castillo2021semi}, the authors learn from a mix of fully annotated images and unannotated ones with auxiliary tasks while~\cite{lucas2021bayesian} designs a custom regularization weighted by the proportion of labeled data to address semi-supervised domain adaptation. In weakly supervised learning, semantic segmentation can be achieved from bounding-boxes~\cite{wang2019weakly}, rough polygons~\cite{daudt2021weakly} or image-level labels~\cite{li2021effectiveness}. Closely related to our kind of supervision,  DISCA~\cite{lenczner2020interactive} and FESTA~\cite{hua2021semantic} leverage point annotations either as user input clicks or partial labels.
 We here use similar  point annotations to add a new segmentation class to a pretrained DNN.

\subsection{Class-incremental semantic segmentation}
The goal of class incremental learning is to modify the output space of DNNs to add label classes. This problem was already considered two decades ago~\cite{bruzzone1999incremental}
and two main pitfalls are  identified. First, like in other class-incremental tasks, it is important to prevent the catastrophic forgetting of the previously learned knowledge. Second, specific to semantic segmentation, the background-shift first tackled in~\cite{cermelli2020modeling} has to be dealt with. Indeed, the new class comes from the background, which causes a discrepancy with the previously learned background class. 
Many methods addressed these two problems by simply storing previous examples~\cite{tasar2019incremental} but more recent works consider this constraint to be too restrictive due to limited storage or security reasons. Hence, we draw inspiration from~\cite{douillard2020podnet,michieli2021continual} which design customized regularization to address the two identified pitfalls.~\cite{douillard2020podnet} proposes distillation losses which enforce statistical matches between the networks. Inspired by few-shot and contrastive learning,~\cite{michieli2021continual} relies on prototypes representing the different semantic classes to distinguish them. 

\section{Methodology}

\subsection{Scenario, baseline and constraints}
A DNN pre-trained for semantic segmentation to detect $N-1$ classes is applied on a new image $I$. The goal is to add a new possible segmentation class and to provide an accurate segmentation map
with respect to these $N$ classes. To this aim, $M$ annotations are provided to make the network interactively learn the $N^{\textit{th}}$ segmentation class. In this work, we follow findings from~\cite{hua2021semantic} and 
use point annotations as they appear to be particularly handy for a potential user. In practice, the last layer of the DNN is modified to increment its output space. As we observed that freezing parts of the network lowered its accuracy, the entire network is then retrained with a cross-entropy (CE) loss on the annotated pixels. However, three problems arise with this simple baseline: (1) \textbf{Catastrophic forgetting} since the new provided labels only belong to the new class; (2) \textbf{Background shift} since the new class was previously classified as background. So, the network has to both learn to detect a new class and to modify the learned representation of the background class; (3) \textbf{Poor class representation} since the network has to learn the new class representation from sparse annotations (i.e. highly incomplete ground-truth).



To address these problems, we notably rely on a pseudo-label strategy and compare different regularizations added to the cross-entropy loss.

Following notation from~\cite{tasar2019incremental}, we refer through the rest of this paper to the previous network with the $N-1$ class prediction as the \textit{memory network} and to the new one as the \textit{updated network}. 
We also refer to the set of semantic classes except the background as the \textit{classes of interest}.

\begin{figure*}[!ht]

\begin{minipage}[t]{\x\linewidth}


\centering{\epsfig{figure=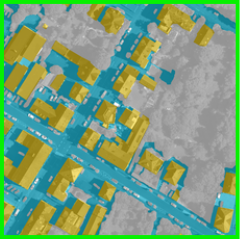,width=\linewidth}}\vspace{.5mm}
\end{minipage}\hfill
\begin{minipage}[t]{\x\linewidth}


\centering{\epsfig{figure=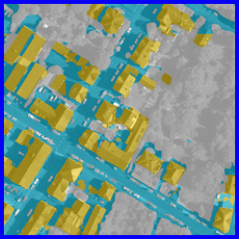,width=\linewidth}}\vspace{.5mm}
\end{minipage}\hfill
\begin{minipage}[t]{\x\linewidth}


\centering{\epsfig{figure=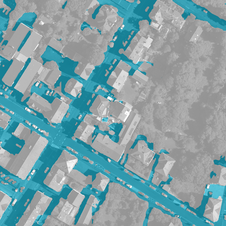,width=\linewidth}}\vspace{.5mm}
\end{minipage}\hfill
\begin{minipage}[t]{\x\linewidth}


\centering{\epsfig{figure=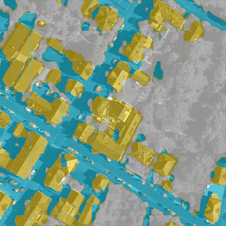,width=\linewidth}}\vspace{.5mm}
\end{minipage}\hfill
\begin{minipage}[t]{\x\linewidth}


\centering{\epsfig{figure=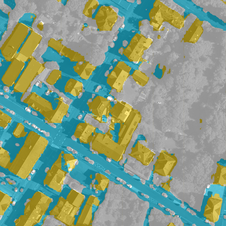,width=\linewidth}}\vspace{.5mm}
\end{minipage}\hfill

\begin{minipage}[t]{\x\linewidth}
\centering\small{Ground-truth}
\end{minipage}\hfill
\begin{minipage}[t]{\x\linewidth}
\centering\small{Control prediction}
\end{minipage}\hfill
\begin{minipage}[t]{\x\linewidth}
\centering\small{Memory prediction} 
\end{minipage}\hfill
\begin{minipage}[t]{\x\linewidth}
\centering\small{Baseline prediction}
\end{minipage}\hfill
\begin{minipage}[t]{\x\linewidth}
\centering\small{SDR prediction}
\end{minipage}\hfill

\caption{Building and road segmentation with 300 building annotations and road pseudo-labels on an example from  Vaihingen compared to ground-truth and control prediction.} 
\label{fig:qualitative_results}
\vspace{-0.2cm}
\end{figure*}

\subsection{Learning from pseudo-labelling}



The prediction of the memory network is used to sample old classes labels. This ensures that the user does not have to provide such old-class annotations while preventing the network from forgetting them. In addition, the interactive user inputs provide both the new class and the background labels. These background labels avoid the aforementioned background-shift issue since background predictions from the memory network could actually belong to the new class. 
\\
To emphasize the N annotations of the new class, only a maximum of N of the most confident old-classes pixels are considered as annotations.


\subsection{Regularization}
In order to make the most of the sparse annotations, we consider the following regularization schemes.

\textbf{DISCA. }Since the new class was previously part of the background, the predictions over the other classes of interest should remain similar. 
To enforce this property, we follow~\cite{lenczner2020interactive} and consider adding a cross entropy regularization term over the pixels which are predicted as belonging to classes of interest by the memory network.
This can be seen as an output-level knowledge distillation loss. 

\textbf{PodNet. }Following PodNet~\cite{douillard2020podnet}, we enforce a statistic match at the encoder level over the channels between the memory network and the updated one with a $L^2$ intermediate-level knowledge distillation loss.

\textbf{SDR. }Inspired by few-shots and contrastive learning, we follow~\cite{michieli2021continual} to regularize the DNN latent space. This aims to reduce forgetting whilst improving the recognition of the new class.
 Concretely, we add a prototype-based regularization at the encoder level of the DNN. Prototypes are vectors representative of each segmentation class and are computed on the fly at each new learning step.
This regularization is composed of three terms: a \textbf{matching term} enforces that the updated prototypes stay close to the previous ones, a \textbf{repulsive term} ensures that the prototypes are far from each other and an \textbf{attracting term} pushes the pixels to remain close to their associated prototype.


\textbf{FESTA. }To deal with sparse annotations for segmentation in remote sensing,~\cite{hua2021semantic} introduces an unsupervised loss that accounts for neighbourhood
structures both in spatial and feature domains as it assumes that nearby pixels share labels.

\section{Experiments}

\subsection{Experimental set-up}

We evaluate our approach on the \textbf{ISPRS Potsdam \& Vaihingen datasets}~\cite{rottensteiner2012isprs} respectively composed of 38 ($6000\times 6000$ pxls, 5cm res.) and 33 (various sizes, 9cm res.) images and on \textbf{SemCity Toulouse}~\cite{roscher2020semcity} composed of 4 images ($3504\times 3452$ pxls, 50 cm res.). We pretrain the network for old class (\textit{road}\footnote{\label{note1}Except when specified otherwise.}) segmentation using dense ground-truth maps on 30 (Potsdam), 26 (Vaihingen) and 2 (Toulouse) images. We fine-tune for new class (\textit{building}\footnotemark[1]) using 300\footnotemark[1] clicked new-class and background annotations simulated from the ground-truth on the remaining data. To compare with a potential upper bound, we consider a control network directly pretrained on dense \{\textit{road, building}\} ground-truth maps (and not fine-tuned). For simplicity, all experiments are performed separately within each dataset (i.e. not in domain adaptation).


The considered metric is the Intersection over Union (IoU) averaged image-wise. Except when we analyze the impact of the number of annotations, we simulate 300 new-class and background annotations. 

For all experiments, we use a LinkNet~\cite{chaurasia2017linknet} architecture trained using Adam optimizer with a learning rate of $10^{-4}$ during
10 pseudo-epochs. Each pseudo-epoch consists in $10\,000$ $256 \times 256$ labeled samples randomly chosen from training data. We infer using a $256 \times 256$ sliding window with an overlap of $50\%$. Due to the stochastic nature of the optimization process and the simulation of the annotations, all experiments are
averaged on 3 runs to obtain statistically significant results.

During fine-tuning, we use an Adam optimizer with a learning rate of $2\cdot10^{-5}$. We fine-tune for 30 steps and select the best performances obtained in the last 15 steps. Each step consists of 10 back-propagation iterations.

\subsection{Approach assessment}


\begin{table}[!h]
\vspace{-.5cm}
\caption{Comparison of the different regularizations. Contr. stands for Control and Bas. for Baseline.}
\vspace{-.2cm}
\small
\setlength{\tabcolsep}{0.35em}
\begin{center}
\begin{tabular}{c|accccc} 

\toprule
 & Contr. & Bas. & FESTA & DISCA & PodNet & SDR\\ 
\midrule
\textit{Pot.}  & 76.4 & 68.7\raisebox{.4ex}{\scriptsize{+- 2.7}} & 66.5\raisebox{.4ex}{\scriptsize{+- 8.4}} & 68.2\raisebox{.4ex}{\scriptsize{+- 4.1}} & 71.7\raisebox{.4ex}{\scriptsize{+- 2.3}} & \textbf{72.0\raisebox{.4ex}{\scriptsize{+- 0.3}}}  \\ 
\textit{Vai.} & 84 & 76.2\raisebox{.4ex}{\scriptsize{+- 2.3}} & 75.1\raisebox{.4ex}{\scriptsize{+- 3.4}} & 74.8\raisebox{.4ex}{\scriptsize{+- 1.8}} & 74.1\raisebox{.4ex}{\scriptsize{+- 5.4}} & \textbf{79.7\raisebox{.4ex}{\scriptsize{+- 1.6}}}   \\ 
\textit{Toul.} & 74 & 63.2\raisebox{.4ex}{\scriptsize{+- 3.2}} & 62.4\raisebox{.4ex}{\scriptsize{+- 1.5}} & 54.0\raisebox{.4ex}{\scriptsize{+- 0.9}} & 65.4\raisebox{.4ex}{\scriptsize{+- 1.0}} & \textbf{67.7\raisebox{.4ex}{\scriptsize{+- 0.6}}}   \\ 

\bottomrule
\end{tabular}
\end{center}
\label{tab:regs}
\end{table}
\vspace{-.3cm}

As indicates Table~\ref{tab:regs}, the DNN is able to learn a new class with clicked annotations. Even without additional regularizations, it reaches an IoU over the three classes of 68.7\% on Potsdam, 76.2\% on Vaihingen and 63.2\% on Toulouse. Appropriate regularization further improves the performances up to 4\%. Indeed, SDR consistently improves the results on the three datasets: 3.3\% on Potsdam, 3.5\% on Vaihingen and 4.5\% on Toulouse. Moreover, SDR tends to stabilize the results compared to the baseline, as testifies the standard deviation measures (e.g. the std on Toulouse is of 0.6 for SDR and of 3.2 for the baseline). As shows Figure~\ref{fig:qualitative_results}, it visually translate into sharper contours, even though the baseline already produces visually accurate results. However, the other regularizations under-perform in this setting in regards to SDR and can even lead to worse results than the baseline on Vaihingen and Toulouse. Hence, the distillation losses and FESTA seem to be less relevant for this problem than the latent space regularization proposed by SDR.

\subsection{Influence of the number of annotations}

\begin{figure}[!h]
  \centering\epsfig{figure=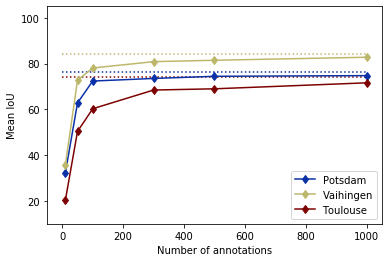, width=.8\linewidth}

  \caption{Metric evolution on the three datasets w.r.t. the number of annotations. Control network performances are in dots.}
  \label{fig:evo}
 \end{figure}
 
To better apprehend our approach, we analyze the influence of the number of annotations over the performances. According to the results of the previous section, we fine-tune the network using SDR regularization. As we can observe on Figure~\ref{fig:evo}, less than one hundred annotations leads to sub-optimal results and the performances then greatly improve with respect to the number of annotations. After 100 annotations for the two ISPRS datasets and 300 annotations for the SemCity dataset, the DNN almost catches up with the control network performances and then stabilizes on a plateau.

\section{Conclusion}
To conclude, we have shown that it is possible to make a DNN learn on the fly new segmentation classes after the initial training with point supervision using pseudo-labelling and relevant regularizations. We have assessed the method efficiency with experiments on three public remote sensing datasets. Moreover, with a sufficient amount of annotations, the performances even almost catch up with a DNN trained on the classes of interest in a fully supervised way.


\begin{spacing}{.95}
\bibliographystyle{IEEEbib}
{
\bibliography{bibli}}

\begin{thebibliography}{10}

\bibitem{castillo2021semi}
J.~Castillo-Navarro et~al.,
\newblock ``{Semi-Supervised Semantic Segmentation in Earth Observation: The
  MiniFrance suite, dataset analysis and multi-task network study},''
\newblock {\em Machine Learning}, pp. 1--36, 2021.

\bibitem{hua2021semantic}
Y.~Hua et~al.,
\newblock ``Semantic segmentation of remote sensing images with sparse
  annotations,''
\newblock {\em Geoscience and Remote Sensing Letters}, vol. 19, pp. 1--5, 2021.

\bibitem{lucas2021bayesian}
B.~Lucas et~al.,
\newblock ``A {B}ayesian-inspired, deep learning-based, semi-supervised domain
  adaptation technique for land cover mapping,''
\newblock {\em Machine Learning}, pp. 1--33, 2021.

\bibitem{wang2019weakly}
Qi~Wang et~al.,
\newblock ``Weakly supervised adversarial domain adaptation for semantic
  segmentation in urban scenes,''
\newblock {\em Trans. on Image Processing}, vol. 28, no. 9, pp. 4376--4386,
  2019.

\bibitem{lenczner2020interactive}
G.~Lenczner et~al.,
\newblock ``Interactive learning for semantic segmentation in earth
  observation,''
\newblock in {\em ECML/PKDD Workshop}, 2020.

\bibitem{daudt2021weakly}
R.~C. Daudt et~al.,
\newblock ``Weakly supervised change detection using guided anisotropic
  diffusion,''
\newblock {\em Machine Learning}, pp. 1--27, 2021.

\bibitem{li2021effectiveness}
Z.~Li et~al.,
\newblock ``On the effectiveness of weakly supervised semantic segmentation for
  building extraction from high-resolution remote sensing imagery,''
\newblock {\em JSTARS}, vol. 14, pp. 3266--3281, 2021.

\bibitem{bruzzone1999incremental}
L.~Bruzzone and D.~Prieto,
\newblock ``An incremental-learning neural network for the classification of
  remote-sensing images,''
\newblock {\em Pattern Recognition Letters}, vol. 20, no. 11-13, pp.
  1241--1248, 1999.

\bibitem{cermelli2020modeling}
F.~Cermelli et~al.,
\newblock ``Modeling the background for incremental learning in semantic
  segmentation,''
\newblock in {\em CVPR}, 2020, pp. 9233--9242.

\bibitem{tasar2019incremental}
O.~Tasar et~al.,
\newblock ``Incremental learning for semantic segmentation of large-scale
  remote sensing data,''
\newblock {\em JSTARS}, vol. 12, no. 9, pp. 3524--3537, 2019.

\bibitem{douillard2020podnet}
A.~Douillard et~al.,
\newblock ``{PodNet}: Pooled outputs distillation for small-tasks incremental
  learning,''
\newblock in {\em ECCV}, 2020, pp. 86--102.

\bibitem{michieli2021continual}
U.~Michieli and P.~Zanuttigh,
\newblock ``Continual semantic segmentation via repulsion-attraction of sparse
  and disentangled latent representations,''
\newblock in {\em CVPR}, 2021, pp. 1114--1124.

\bibitem{rottensteiner2012isprs}
F.~Rottensteiner et~al.,
\newblock ``The {ISPRS} benchmark on urban object classification and {3D}
  building reconstruction,''
\newblock in {\em ISPRS Annals}, 2012, vol. 1, no. 1, pp. 293--298.

\bibitem{roscher2020semcity}
R.~Roscher et~al.,
\newblock ``{SemCity Toulouse}: A benchmark for building instance segmentation
  in satellite images,''
\newblock {\em ISPRS Annals}, vol. 5, pp. 109--116, 2020.

\bibitem{chaurasia2017linknet}
A.~Chaurasia and E.~Culurciello,
\newblock ``{LinkNet: Exploiting} encoder representations for efficient
  semantic segmentation,''
\newblock in {\em VCIP}, 2017, pp. 1--4.

\end{thebibliography}
\end{spacing}
\end{document}